\documentclass{article}
\usepackage{spconf,amsmath,graphicx}
\graphicspath{{../img/}}

\title{Cross-lingual and Multilingual Speech Emotion Recognition on English and French}
\name{Michael Neumann, Ngoc Thang Vu}
\address{University of Stuttgart, Germany\\
	      \{michael.neumann$|$thang.vu\}@ims.uni-stuttgart.de}
\begin{document}
%
\maketitle
\begin{abstract}
Research on multilingual speech emotion recognition faces the problem that most available speech corpora differ from each other in important ways, such as annotation methods or interaction scenarios. These inconsistencies complicate building a multilingual system.
We present results for cross-lingual and multilingual emotion recognition on English and French speech data with similar characteristics in terms of interaction (human-human conversations). Further, we explore the possibility of fine-tuning a pre-trained cross-lingual model with only a small number of samples from the target language, which is of great interest for low-resource languages.
To gain more insights in what is learned by the deployed convolutional neural network, we perform an analysis on the attention mechanism inside the network.
\end{abstract}
\keywords{Speech Emotion Recognition, Multilingual, Cross-lingual, CNN, Attention}

\section{Introduction}

The common approach to automatic emotion recognition is to train and test a classifier on one annotated (mostly mono-lingual) corpus, either by subdividing the data into train, validation and test sets or by means of cross-validation. This way, the system is highly specialized with respect to a number of factors, such as the speaker group, the recording situation, the language, and the type of speech (spontaneous or acted). Further, no conclusions can be drawn to what extend such a system can generalize across different interaction scenarios and languages.
For this reason, we investigate cross-lingual and multilingual speech emotion recognition, as a step towards language-independent emotion recognition in natural speech.

In addition to the aforementioned reasons, cross-lingual classification can possibly facilitate emotion recognition for scenarios with no or only a small amount of annotated data in the target language, which we refer to as low-resource setting. 

Various cross-corpus analyses have been conducted in recent years~\cite{schuller2010cross, lefter2010emotion, eyben2010cross, schuller2011using, schuller2011selecting}. 
In an extensive study with six corpora,~\cite{schuller2010cross} examined many different combinations of corpora as training set, without focusing on one certain aspect of the data (e.g. different language or different interaction scenario). Although this study gives an overall impression on the performance of cross-corpus emotion recognition, it makes the interpretation of results difficult because it is not clear which factors have what kind of impact.
Focusing on cross-language emotion recognition,~\cite{feraru2015cross} presented a comprehensive overview using 8 languages from 4 language families and showed that cross-language emotion recognition is feasible, but with notably lower accuracy than mono-lingual recognition.

An approach to multilingual emotion classification using language identification and model selection is presented in~\cite{sagha2016enhancing}. In contrast to this work where language-dependent models are trained and then selected accordingly, we examine the performance of one model trained on multiple languages. Another strategy to combine two languages for emotion recognition, described in~\cite{chiou2014speech}, is to apply histogram equalization to remove cross-language variability. In~\cite{jeon2013preliminary}, the authors compare automatic cross-lingual recognition with human perception of emotion.

Concerning classification performance it is difficult to compare to related research in this field, because there are no standards regarding several factors, including the number of classes, the division of corpora into train and test sets, the underlying emotion concepts (categorical emotions or continuous arousal/valence dimensions).
Hence, we cannot discuss state-of-the-art performance in this study, because the aforementioned works differ in at least one of these respects, mostly in the number of classes, the utilized databases or the mapping between continuous and discrete annotations.
The focus of the present research is on multilingual and cross-lingual speech emotion recognition compared to mono-lingual baselines trained on the respective corpora, as well as on an analysis of the attention mechanism used in the recognition system. We show that multilingual emotion recognition is feasible without adaptation to the language and present promising results for cross-lingual training followed by fine-tuning on the target language.

\section{Model Architecture}
\begin{figure}
\center
\includegraphics[trim={60 440 60 65}, clip, width=.5\textwidth]{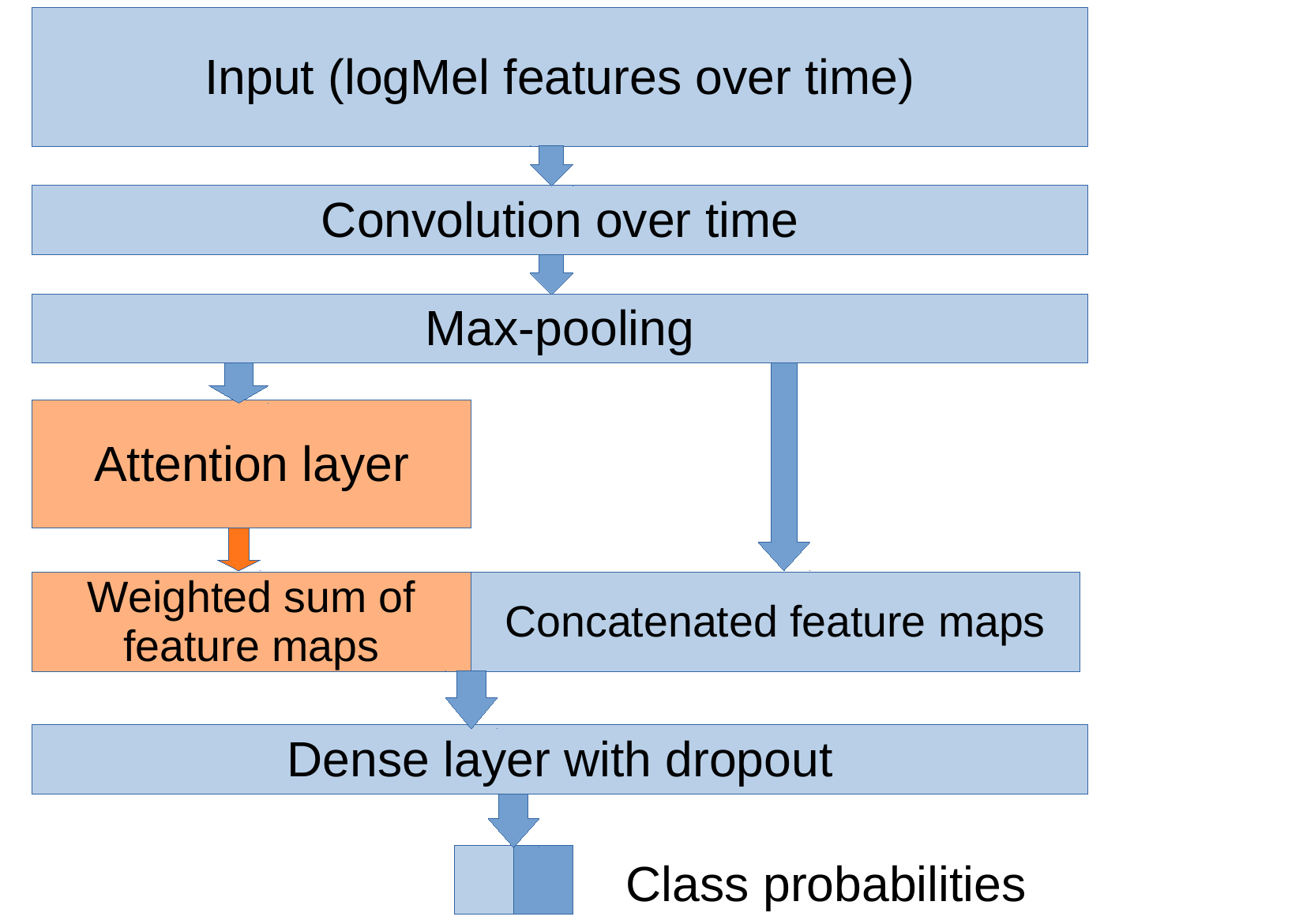}
\caption{Model topology.}
\label{fig:acnn}
\end{figure}

For this study we train an attentive convolutional neural network (ACNN) for binary classification of arousal and valence in speech. The model architecture is mainly adopted from~\cite{neumann2017attentive} and adjusted to this task and the cross-lingual setting. Figure~\ref{fig:acnn} depicts the network topology schematically. As input features to the ACNN, 26 logMel filter-banks are extracted frame-wise from the segmented speech signal (frame size of 25ms and a window shift of 10ms). The input has fixed length of 7.5s, shorter utterances are padded with zeros at the end. The convolution kernels span all 26 features (\mbox{1-D} convolution over time).
The output from the max-pooling layer is fed into an attention layer which computes a weighted sum of the information extracted from different parts of input. The input to the fully connected softmax layer at the end is the concatenation of attention vector and the feature maps from the pooling layer.  

The attention layer computes attention weights $\alpha_i$ over all feature maps for each time step $i$.
Equation~\ref{eq:att} shows the computation of these attention weights $\alpha_{i} $ for an input sequence $x$ consisting of vectors $x_i$, where $f(x)= W^T x$, with $W$ being a trainable parameter.
\begin{equation}
\alpha_i = \frac{exp(f(x_i))}{\sum_{j} {exp(f(x_j))}}
\label{eq:att}
\end{equation}

The output of the attention layer, $attentive\_x$, is the weighted sum of the input sequence.
\begin{equation}
attentive\_x= \sum_{i} \alpha_{i} x_i
\end{equation}

The intuition behind using an attention mechanism for emotion recognition is that emotional information is distributed differently over the signal. 
Therefore, we want to first weight the information extracted from different pieces of the input and then combine them in a weighted sum.

\section{English and French Emotional Speech}
We use two corpora of emotional speech which are frequently used and freely available. The main criterion for selecting the data was that the corpora contain the same type of speech in terms of conversation type (human-human) and naturalness.

The interactive emotional dyadic motion capture database (\textbf{IEMOCAP})~\cite{busso2008iemocap} is a multimodal database of English dyadic conversations containing both fixed speech (scripted dialogs) and free, spontaneous speech (improvised dialogs given a certain scenario and topic). The speakers are professional actors. IEMOCAP is annotated on turn level in two ways, with categorical emotion labels (such as 'anger', 'happiness', 'sadness') and with 5-point scales on the dimensions valence, arousal and dominance (1 - low/negative, 5 - high/positive). The corpus contains 10,039 utterances.

\textbf{Recola}~\cite{ringeval2013introducing} is a multimodal database of French speech consisting of dyadic conversations during a video conference where participants had to solve a collaborative task. From 46 speakers in total, we use the freely available portion of 23 speakers in this study, consisting of 1,308 utterances. Recola is annotated with continuous labels for arousal and valence in the range [-1, 1] on a 40ms rate. Annotation was done with ANNEMO~\cite{ringeval2013introducing}, a tool similar to Feeltrace~\cite{cowie2000feeltrace}. Since we are interested in recognition of emotions on utterance level, we calculated the mean of all values for one turn, and then took the average across all annotators as the final label.

To be able to train a model on several corpora, the problem of different annotation schemes has to be overcome. We decide to focus on a binary classification task of arousal (low/high) and valence (negative/positive).\footnote{class distribution IEMOCAP: arousal - 3,121 low, 6,918 high; valence - 3,421 neg., 6,618 pos. | Recola: arousal - 520 low, 788 high; valence - 241 neg., 1,067 pos.} The mapping of original annotations to a binary scheme is shown in Table~\ref{tab:mapping}. 
\begin{table}[htbp]
\centering
\begin{tabular}{l|l|l}
 & \textbf{Low/Negative} & \textbf{High/Positive} \\ \hline \hline
\textbf{IEMOCAP} & range [1, 2.5] & range (2.5, 5] \\ \hline
\textbf{Recola} & range [-1, 0] & range (0, 1] \\
\end{tabular}
\caption{Mappings to binary arousal/valence classes.}
\label{tab:mapping}
\end{table}

\section{Experimental Setup}
We conduct the following four experiments: (a) mono-lingual (as baseline), (b) multilingual (merge Recola and IEMOCAP for training), (c) cross-lingual (train on one corpus, test on the other one), and (d) fine-tuning of a model trained in (c) in a simulated low-resource setting.

For (a) mono-lingual and (b) multilingual experiments we apply cross validation (CV) because there are no predefined train and test splits for these datasets. The IEMOCAP data consists of five sessions with one male and one female speaker each. We take data from four sessions to construct training and development sets and use the remaining session for testing, resulting in 5-fold CV. For Recola, we construct manually five splits so that they are balanced with respect to number of speakers and sex. This way, we ensure speaker-independent training (in contrast to random sampling). 

The evaluation of (c) cross-lingual training is more straightforward, we take all data of one language as training set and all samples of the respective other as test set.
For~(d) fine-tuning (FT) in the simulated low-resource setup we take trained models from (c) as starting point. The model is then refined using 100 randomly selected samples from the target language for each CV split. Consequently, only 500 samples of the target language are used in total for FT.

In order to observe variations in the results due to non-deterministic operations on the GPU, we run all experiments five times and report the means.

\subsubsection*{Hyper-parameters}
The ACNN model is implemented with the Tensorflow library \cite{abadi2016tensorflow}. We apply stochastic gradient descent with an adaptive learning rate (Adam~\cite{kingma2015adam}) for training.
The systems hyper-parameters are the following: 200 kernels with a size of 26x10 in the convolutional layer (spanning all 26 logMel filter-banks); a mini-batch size of~32; and a pool size of~30 for max-pooling. For regularization we apply dropout (\cite{srivastava2014dropout}) to the last hidden layer with a dropout rate of 0.5. We run training for 50 epochs in all experiments except for fine-tuning where the pre-trained models are refined with only 10 epochs.

The kernel width of 10 and the pool size of 30 were selected by tuning the mono-lingual models on the development set with a number of different parameter combinations.
These relatively high values appear reasonable considering that emotions in speech are determined as long-term information. The kernel size of 10 corresponds to 100ms from the input signal. A large amount of overlap in the feature maps explains the large pooling window.

\section{Results}
The performance measure used throughout all experiments is unweighted average recall (UAR, i.e. the average of the recall for each class). This measure best reflects the overall accuracy when the dataset is imbalanced with respect to the number of samples per class.
The results are presented in Table~\ref{tab:results}.

\begin{table}[htbp]
\centering
\begin{tabular}{l|c|c|c|c}
 & \multicolumn{2}{c|}{IEMOCAP} & \multicolumn{2}{c}{Recola} \\
 & \multicolumn{2}{c|}{(English)} & \multicolumn{2}{c}{(French)} \\
 & Arousal & Valence & Arousal & Valence \\ \hline
mono-lingual & 68.09 & \textbf{62.33} & 60.77 & \textbf{52.30} \\ \hline
multilingual & \textbf{70.06} & 61.73 & 62.51 & 49.33 \\ \hline \hline
cross-lingual & 59.32 & 49.08 & 61.27 & 47.52 \\ \hline 
CL + FT & 67.03 & 50.42 & \textbf{63.07} & 49.81 \\
\end{tabular}
\caption{Results as unweighted average recall (UAR), cross-lingual: only trained on source language, CL + FT: pre-trained on source language and fine-tuned on 500 samples from target lanugage (CL - cross-lingual, FT - fine-tuning).}
\label{tab:results}
\end{table}
The mono-lingual baselines for both IEMOCAP and Recola show that the prediction of valence is more difficult than arousal. This finding is in line with~\cite{eyben2010cross, feraru2015cross, trigeorgis2016adieu}. The performance for Recola is notably lower than for IEMOCAP. This is only partially due to the small size of Recola (1,308 samples). Using only 1,308 samples from IEMOCAP in comparison still leads to better results for English (68.20\% arousal and 58.77\% valence). Another possible reason is that the French data is highly imbalanced for valence (UAR of 52.30\% is only slightly better than chance). 

With multilingual training we want to investigate the effect of merging the two corpora and find out whether multilingual speech emotion recognition is possible without performance loss. The results show that we are able to use a system trained on both languages and achieve similar performance compared to the baselines.
For arousal prediction, the additional training data even improves performance, whereas we observe a decrease in performance for valence.
These findings are a first evidence that multilingual speech emotion recognition is viable without further adaptation.

Cross-lingual training is useful in cases where no or only little training data in the target language is available. We therefore examine the performance of the system when trained on one language and tested on the other (and vice versa), given the same type of speech (human-human interaction). The results in Table~\ref{tab:results} show that cross-lingual training works to some extend for arousal but not for valence prediction. For arousal, the performance drops notably for IEMOCAP (trained on Recola) compared to the mono-lingual baseline, achieving 59.32\% UAR. For Recola (trained on IEMOCAP) it remains stable (60.77\% mono-lingual, 61.27\% cross-lingual). For valence, both results are below chance, suggesting that valence prediction might be more language-dependent than predicting arousal.

Fine-tuning the cross-lingual model with 10 training epochs on 500 samples from the target language produces promising results for arousal prediction. For IEMOCAP, the performance comes close to the baseline and for Recola, it is notably higher than the baseline. Again, the performance for valence remains approximately at chance level. 

In summary, these results show that cross-lingual training can set a useful baseline. Especially for a target language with a small amount of annotated data, 
training a cross-lingual model and then fine-tuning it on the available target data appears to be a reasonable approach.

\section{Analysis of attention weights}
\begin{figure}
\includegraphics[trim={0 0 0 35}, clip, width=.51\textwidth]{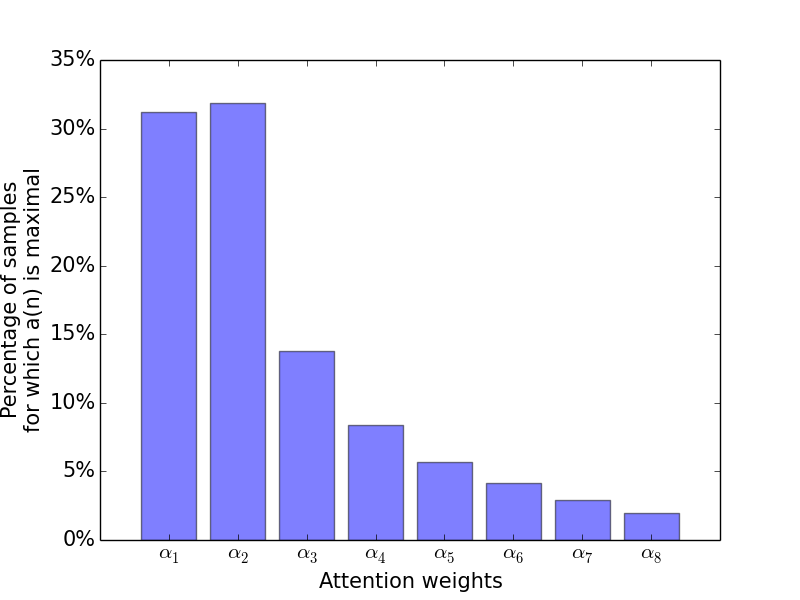}
\caption{Distribution of attention over time for arousal prediction on IEMOCAP.}
\label{fig:att_weights}
\end{figure}

\begin{figure}
\includegraphics[trim={0 0 0 35}, clip, width=.51\textwidth]{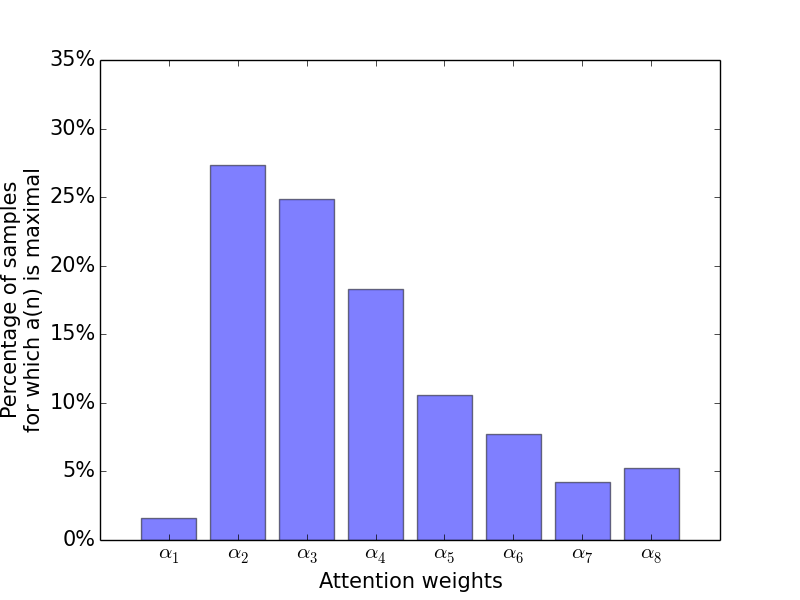}
\caption{Distribution of attention over time for arousal prediction on Recola.}
\label{fig:att_weights2}
\end{figure}

To gain more insights about which parts of the input are important for classification, we analyze the attention weights $\alpha_{i}$ from the attention layer after the last training epoch. 
We focus on the mono-lingual baseline experiments in arousal prediction. For each training sample, we output its attention weights and identify the maximum weight, i.e. the segment which appears to be most salient for this sample.
Figures~\ref{fig:att_weights} and~\ref{fig:att_weights2} show for every attention weight $\alpha_1$ to $\alpha_8$ the particular proportion of training samples for which this $\alpha_i$ yielded the maximum value. For example in Figure~\ref{fig:att_weights}, for 31.2\% of training samples, $\alpha_1$ was highest, hence the first segment of the input to the attention layer is considered most salient. The number of attention weights corresponds to the output vector of the max-pooling layer and therefore depends on input signal length, kernel size and pool size. Figure~\ref{fig:att_weights} shows this distribution for the English data and Figure~\ref{fig:att_weights2} for French.



From Figure~\ref{fig:att_weights} it can be observed that for a large majority of samples the attention lies at the beginning of the input. This finding is in line with the observation in~\cite{neumann2017attentive} that a short snippet from the beginning of an utterance can be sufficient for a prediction in many cases.
In addition to depicting the maximum attention weights, we took a closer look at the actual values of the maximum and the second highest weight to find out more about the weight distribution. Note, that the weights $\alpha_1$ to $\alpha_8$ sum up to 1.0 for every sample. For the English training data we found that for 73.1\% of all samples the difference between highest and second highest attention weight is greater than 0.5. This means, for the majority of data one segment is weighted much higher than all others. 

For the French data, the picture looks a bit different. Figure~\ref{fig:att_weights2} reveals that $\alpha_2$ to $\alpha_4$ yield the maximum weight for a large proportion of data. Apart from $\alpha_1$, the distribution exhibits similar characteristics as in Figure~\ref{fig:att_weights}, that the beginning of the input is much more often considered important than the end. Our first hypothesis to explain the low rate for $\alpha_1$ was that many samples contain silence at the beginning. However, using voice activity detection, we found that most signals contain speech straight from the beginning. Hence, further analysis is necessary to explain this difference.
In the Recola dataset the difference between highest and second highest attention weights is only for 5\% of training samples greater than 0.5. 
This overall flatter distribution suggests that it is more difficult to learn meaningful attention weights for the French data compared to English.

 
To conclude this analysis, we have found notable differences in the attention mechanism between the two datasets. But it is difficult to draw final conclusions about the languages themselves because the corpora are not recorded under same conditions (especially the underlying task for the participants). Hence, these findings point towards language-dependent characteristics in emotional speech, but are potentially skewed by language-independent factors such as recording situation or the lexical content of the conversations.


\section{Conclusion}
We presented results for binary arousal/valence classification using cross-lingual and multilingual training. We have shown that multilingual classification of emotions in speech is possible and can even enhance results for arousal prediction. This can be regarded as a valuable finding for research on code-switching speech.
Further, we have shown that a model trained on a source language and fine-tuned with only a small number of samples from the target language can produce sound results for arousal prediction, whereas valence prediction appears to be more sensitive to cross-lingual training. These findings are potentially interesting for emotion research on low-resource languages.

\section{Acknowledgement}
This work was funded by the German Science Foundation (DFG), Sonderforschungsbereich 732 Incremental Specification in Context, Project A8, at the University of Stuttgart.

\vfill\pagebreak

\bibliographystyle{IEEEbib}
\bibliography{../bibliography}

\end{document}